\documentclass[runningheads]{llncs}
\usepackage{graphicx}

\usepackage{comment}
\usepackage{amsmath,amssymb} %
\usepackage{color}

\usepackage{url}
\usepackage{rotating}      %
\usepackage[utf8]{inputenc} %
\usepackage[T1]{fontenc}    %
\usepackage{hyperref}       %
\usepackage{url}            %
\usepackage{booktabs}       %
\usepackage{amsfonts}       %
\usepackage{nicefrac}       %
\usepackage{microtype}      %
\usepackage{color,xcolor}
\usepackage{enumitem}
\usepackage{bm}
\usepackage{multirow}
\usepackage{bbm}
\usepackage{algorithm} 
\usepackage{makecell}
\usepackage{tikz}
\usetikzlibrary{spy}

\newcommand{\sidecap}[1]{ {\begin{sideways}\parbox{1.6cm}{\centering #1}\end{sideways}} }

\makeatletter 
  \newcommand\figcaption{\def\@captype{figure}\caption} 
  \newcommand\tabcaption{\def\@captype{table}\caption} 
\makeatother

\usepackage{graphicx}
\graphicspath{{images/}}

\def\bmx{\bm{x}}
\def\bmy{\bm{y}}
\def\bmz{\bm{z}}

\def\calI{{\mathcal{I}}}

\def\qut#1{\left(#1\right)}
\def\qutb#1{\left[#1\right]}
\def\qutl#1{\mathopen{}\left(#1\right)\mathclose{}}
\def\bmth{\bm{\theta}}
\def\bmc{\bm{c}}
\def\bmsk#1{\bm{s}^{(#1)}}

\def\sk#1{{s}^{(#1)}}
\def\hk#1{{h}^{(#1)}}
\def\calI{{\mathcal{I}}}

\usepackage{cite}

\begin{document}
\pagestyle{headings}
\mainmatter
\def\ECCVSubNumber{100}  %

\title{Learning to segment microscopy images with lazy labels} %

\titlerunning{Learning to segment microscopy images with lazy labels}
\author{Rihuan Ke\inst{1} \orcidID{0000-0002-8086-3441} 
\and
Aur\'elie Bugeau \inst{2} \orcidID{0000-0002-4858-4944} 
\and
Nicolas Papadakis \inst{3} \orcidID{0000-0002-5414-4902} 
\and
Peter Schuetz \inst{4} \and 
Carola-Bibiane Sch{\"o}nlieb \orcidID{0000-0003-0099-6306}
\inst{1}}
\authorrunning{R. Ke et al.}
\institute{
DAMTP, University of Cambridge, Cambridge CB3 0WA, UK
\email{rk621@cam.ac.uk}
\and
LaBR, University of Bordeaux, UMR 5800, F-33400 Talence, France
\and 
CNRS, University of Bordeaux, IMB, UMR 5251, F-33400 Talence, France
\and
Unilever R\&D Colworth, MK44 1LQ, UK
}
\maketitle

\begin{abstract}
The need for labour intensive pixel-wise annotation
is a major limitation of many fully supervised learning methods for segmenting bioimages that can contain numerous object instances with thin separations. 
In this paper, we introduce a deep convolutional neural network for microscopy image segmentation. Annotation issues are
circumvented by letting the network being trainable on coarse labels
combined with only a very small number of images with pixel-wise annotations. We call this new labelling strategy `lazy' labels.
Image segmentation is  stratified into three connected tasks: rough inner region detection, object separation and pixel-wise segmentation. These tasks are learned in an end-to-end multi-task learning framework. 
The method is demonstrated on two microscopy datasets, where we show that the model gives accurate segmentation results even if exact boundary labels are missing for a majority of annotated data. 
It brings more flexibility and efficiency for training deep neural networks that are data hungry and is applicable to biomedical images with poor contrast at the object boundaries or with diverse textures and repeated patterns. 
\end{abstract}

\section{Introduction}

Image segmentation is a crucial step in many microscopy image analysis problems. 
It has been an active research field in the past decades. 
Deep learning approaches play an increasingly important role and have become state-of-the-art in various segmentation tasks \cite{huang2018weakly, khoreva2017simple, tsutsui2018minimizing, ghosh2018stacked, litjens2017survey}.
However, the segmentation of microscopy images is very challenging not only due to the fact that these images are often of low contrast with complex instance structures, but also because of the difficulty in obtaining ground truth pixel-wise annotations \cite{hirsch2020auxiliary, bajcsy2020approaches} which hinders the applications of recent powerful but data-hungry deep learning techniques.

In this paper, we propose a simple yet effective multi-task learning approach for microscopy image segmentation. We address the problem of finding segmentation with accurate object boundaries from mainly rough labels. The labels are all pixel-wise and contain considerable information about individual objects, but they are created relatively easily. 
The method is different from pseudo labelling (PL) approaches, which generate fake training segmentation masks from coarse labels and may induce a bias in the masks for microscopy data.

\begin{figure}[ht]
  \centering
\includegraphics[width=0.8\linewidth, trim=0 0 50 0 ,clip]{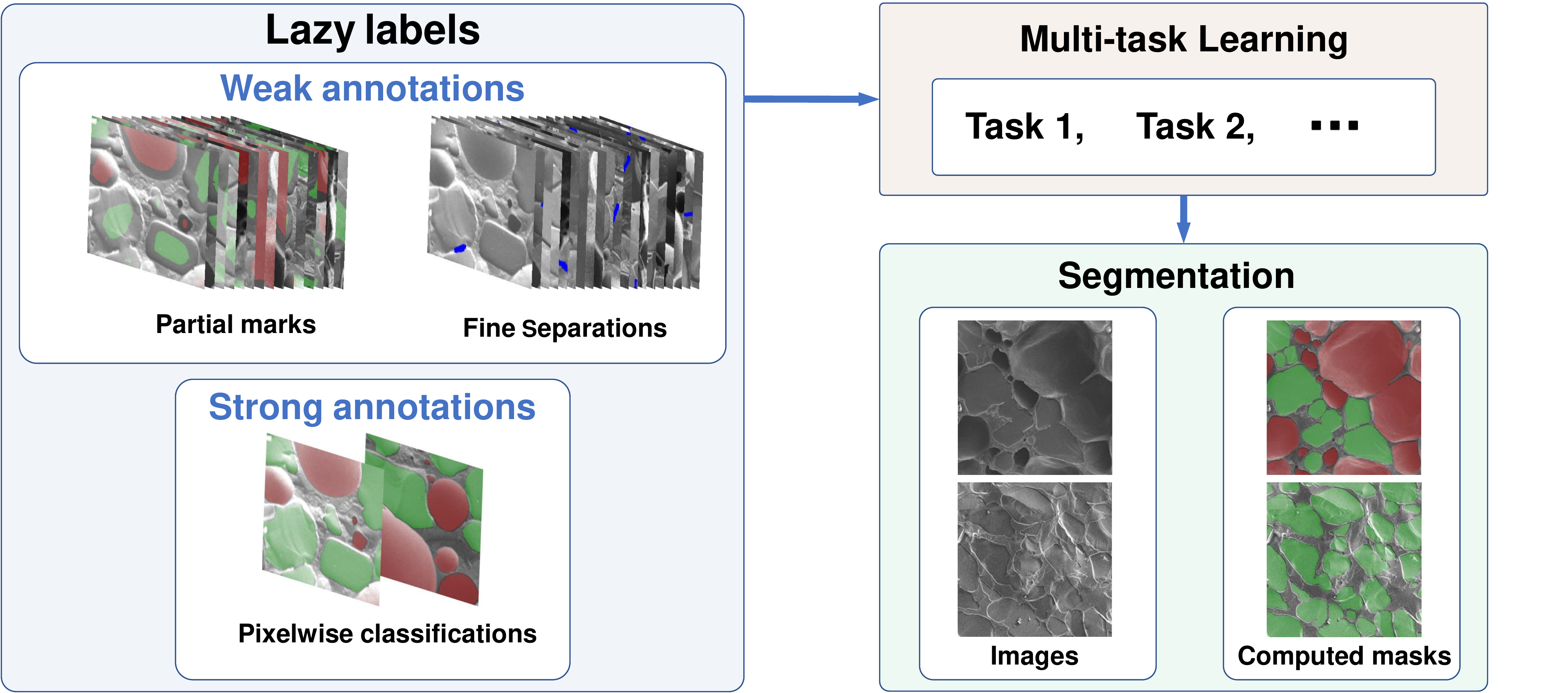}
  \caption{Multi-task learning for image segmentation with lazy labels. The figure uses Scanning Electron Microscopy (SEM) images of food microstructures as an example and demonstrates a segmentation problem of three classes, namely air bubbles (green), ice crystals (red) and background respectively.
Most of the training data are weak annotations containing (i) partial  marks of ice crystals and/or air bubbles  instances and (ii) fine separation marks of boundaries shared by different instances.
Only a \textit{few} strongly annotated images are used. %
On the bottom right SEM images and their corresponding segmentation outputs from the learned multi-task model are shown. 
}\label{fig:icecreamimages}
\end{figure}

To circumvent the need for a massive set of ground truth segmentation masks, we rather develop a segmentation approach that we split   
into three relevant tasks: detection, separation and segmentation (cf. Figure \ref{fig:icecreamimages}). 
Doing so, we obtain a weakly supervised learning approach that is trained with what we call "lazy" labels. These lazy labels contain a lot of coarse annotations of class instances, together with a few accurately annotated images that can be obtained from the coarse labels in a semi-automated way.
Contrary to PL approaches, only a very limited number of accurate annotation are considered. In the following, we will refer to weak (resp. strong) annotations for coarse (resp. accurate) labels and denote them as WL (resp. SL).

We reformulate the segmentation problem into several more tractable tasks that are trainable on less expensive annotations, and therefore reduce the overall annotation cost. 
The \textit{first task} detects and classifies each object by roughly determining its inner region with an under-segmentation mask. Instance counting can be obtained as a by-product of this task.
As the main objective is instance detection, exact labels for the whole object or its boundary are not necessary at this stage. 
We use instead weakly annotated images in which a rough region inside each object is marked, cf. the most top left part of Figure  \ref{fig:icecreamimages}.
For segmentation problems with a dense population of instances, such as the food components (see e.g., Figure \ref{fig:icecreamimages}), cells \cite{guerrero2018multiclass,ronneberger2015u}, glandular tissue, or people in a crowd \cite{wang2018repulsion}, separating objects sharing a common boundary is a well known challenge. We can optionally perform a \textit{second task} that focuses on the separation of instances that are connected without a clear boundary dividing them.
Also for this task we rely on WL to reduce the burden of manual annotations: touching interfaces are specified with rough scribbles, cf. top left part of Figure \ref{fig:icecreamimages}. Note that this task is suitable for applications with instances that are occasionally connected without clear boundaries. One can alternatively choose to have fewer labelled samples in this task if the annotation cost per sample is higher. The \textit{third task} finally tackles  pixel-wise classification of  instances. It requires strong annotations that are accurate up to the object boundaries. Thanks to the information brought by weak annotations, we here just need a very small set of accurate segmentation masks, cf. bottom left part of Figure \ref{fig:icecreamimages}. To that end, we  propose to refine some of the coarse labels resulting from task 1 using a semi-automatic segmentation method which requires additional manual intervention. 

The three tasks are handled by a single deep neural network and are jointly optimized using a cross entropy loss. 
In this work we use a network architecture inspired by  U-net \cite{ronneberger2015u} which is widely used for segmenting objects in microscopy images.  
While all three tasks share the same contracting path, we introduce a new multi-task block for the expansive path.
The network has three outputs and is fed with a combination of WL and SL described above. 
Obtaining accurate segmentation labels for training is usually a hard and time consuming task. We here demonstrate that having exact labels for a small subset of the whole training set does not degrade training performances. We evaluate the proposed approach on two microscopy image datasets, namely the segmentation of SEM images of food microstructure and stained histology images of glandular tissues.

The contributions of the paper are threefold.
(1) We propose a decomposition of the segmentation problems into three tasks and a corresponding user friendly labelling strategy.
(2) We develop a simple and effective multi-task learning framework that learns directly from the coarse and strong manual labels and is trained end-to-end. (3) Our approach outperforms the pseudo label approaches on the microscopy image segmentation problems being considered.

\section{Related Work}%

In image segmentation problems, one needs to classify an image at pixel level.
It is a vast topic with a diversity of algorithms and applications being considered,
including traditional unsupervised methods like $k$-means clustering \cite{macqueen1967some} that splits the image into homogeneous
regions according to image low level features,
curve evolution based methods like snakes \cite{caselles1997geodesic},
graph-cut based methods \cite{rother2004grabcut, krasowski2017neuron, boykov2006graph}, just to name a few.
Interactive approaches like snakes or Grabcut enable getting involved users' knowledge by means of
initializing regions or putting constraints on the segmentation results.
For biological imaging, the applications of biological prior knowledge, such as shape statistics
\cite{heimann2009statistical}, semantic information
\cite{krasowski2017neuron} and atlas
\cite{ciofolo2009atlas}, is effective for automatic segmentation approaches. 

\subsection{Deep neural networks for segmentation}
In the last years, numerous deep convolutional neural network (DCNN) approaches have been developed for segmenting  complex images, especially in the semantic setting.  In this work, we rely more specifically on fully convolutional networks (FCN) \cite{long2015fully}, that 
replace the last few fully connected layers of a conventional classification network by up-sampling layers and convolutional layers, to preserve  spatial  information.
FCNs have many variants for semantic segmentation.
The DeepLab \cite{chen2018deeplab} uses a technique called atrous convolution to handle spatial information together with a %
fully connected conditional random field (CRF) \cite{chen2014semantic} for refining the segmentation results.
Fully connected CRF can be used as post-processing %
or can be integrated into the network architecture, allowing for end-to-end training \cite{zheng2015conditional}.

One type of FCNs commonly used in  microscopy and biomedical image segmentation are  encoder-decoder networks \cite{badrinarayanan2017segnet,ronneberger2015u}.
They have multiple up-sampling layers for better localizing boundary details.
One of the most well-known models is the U-net~\cite{ronneberger2015u}. %
It is a fully convolutional network made of a contracting path, which brings the input images into very low resolution features with a sequence of down-sampling layers,  and an  expansive path that has an equal amount of up-sampling layers. At each resolution scale, the features on the contracting path are merged with the corresponding up-sampled layers via long skip connections  to recover detailed structural information, e.g., boundaries of cells, after down-sampling.

\subsection{Weakly supervised learning and multi-task learning}
Standard supervision for semantic segmentation relies on a set of image and ground truth segmentation pairs. The learning process contains an optimization step that minimizes the distance between the outputs and the ground truths.  
There has been a growing interest in weakly supervised learning,  motivated by the heavy cost of pixel-level annotation needed for fully supervised methods. 
Weakly supervised learning uses weak annotations such as image-level labels \cite{huang2018weakly, papandreou2015weakly,lee2019ficklenet, mlynarski2018deep,playout2019novel, zhou2019weakly, shin2019joint}, bounding boxes\cite{khoreva2017simple, shah2018ms}, scribbles \cite{lin2016scribblesup} and points \cite{bearman2016s}.

Many weakly supervised deep learning methods for segmentation are built on top of a classification network. The training of such networks may be realized using  segmentation masks explicitly generated from weak annotations
\cite{wei2017stc,khoreva2017simple, lee2019ficklenet, wei2018revisiting, tsutsui2018minimizing}.
The segmentation masks can be improved recursively, which involves several rounds of training of the segmentation network \cite{wei2017stc, jing2018coarse,ezhov2018coarse}. Composite losses from some predefined guiding principles are also proposed as supervision from the weak signals  \cite{kolesnikov2016seed,kolesnikov2016seed, sun2019saliency}.

Multi-task techniques aim to boost the segmentation performance via learning jointly from several relevant tasks. 
Tailored to the problems and the individual tasks of interest, deep convolutional networks have been designed, for example, the stacked U-net for extracting roads from satellite imagery \cite{sun2018stacked}, the two stage 3D U-net framework for $3$D CT and MR data segmentation \cite{wang2018two}, encoder-decoder networks for depth regression, semantic and instance segmentation \cite{kendall2018multi}, or the cascade multi-task network for the segmentation of building footprint \cite{bischke2019multi}. 

\paragraph{Segmentation of microscopy and biomedical images.}
Various multi-task deep learning methods have been developed for processing microscopy images and biomedical images. 
An image level lesion detection task \cite{playout2019novel} is investigated for the segmentation of retinal red/bright lesions. 
The work \cite{mlynarski2018deep} considers to jointly segment and classify brain tumours. 
A deep learning model is developed in  \cite{zhou2019high} to simultaneously predict the segmentation maps and contour maps for pelvic CT images. 
In \cite{hirsch2020auxiliary}, an auxiliary task that predicts centre point vectors for nuclei segmentation in 3D microscopy images is proposed. Denoising tasks, which aims to improve the image quality, can also be  integrated for better microscopy image segmentation \cite{buchholz2020denoiseg}. 

In this work, the learning is carried out in a weakly supervised fashion with weak labels from closely related tasks.  
Nevertheless, the proposed method exploits cheap and coarse pixel-wise labels instead of very sparse image level annotations
and is more specialized in distinguishing the different object instances and clarifying their boundaries in microscopy images. The proposed method is completely data-driven and it significantly reduces the annotation cost needed by standard supervision. 
We aim at obtaining segmentation with accurate object boundaries from mainly coarse pixel-wise labels.

\section{Multi-task learning framework}\label{sec:model}

The objective of fully supervised learning for segmentation is to approximate the conditional probability distribution of the segmentation mask given the image. Let {\small $\bmsk{3}$} be the ground truth segmentation mask and $I$ be the image, then the  segmentation task aims to estimate {\small $p(\bmsk{3} \mid I)$} based on a set of sample images {\small $\calI = \{I_1, I_2, \cdots, I_n\}$} and the corresponding labels {\small $\{ \bmsk{3}_1, \bmsk{3}_2, \cdots, \bmsk{3}_n \}$}. The set {\small $\calI$} is randomly drawn from an unknown distribution.
In our setting, having the whole set of segmentation labels {\small $\{ \bmsk{3}_i \}_{1,\cdots,n}$} is impractical, and we introduce two auxiliary tasks for which the labels can be more easily generated to achieve an overall small cost on labelling.

For a given image {\small $I\in \calI$},  we denote as {\small $\bmsk{1}$} the rough instance detection mask,  %
and {\small $\bmsk{2}$} a map containing some interfaces shared by touching objects. All labels $\bmsk{1},\bmsk{2},\bmsk{3}$ are represented in one-hot vectors.  
For the first task, the contours of the objects are not treated carefully, resulting in a coarse label mask {\small $\bmsk{1}$} that misses most of the boundary pixels, cf left of Figure  \ref{fig:icecreamimages}.
In the second task, the separation mask {\small $\bmsk{2}$} only specifies connected objects without clear boundaries rather than their whole contours. 
Let {\small $\calI_k \subset{\calI}$}  denote the subset of images labelled for task $k$ $\qutl{k=1,2,3}$.  
As we collect a different amount of annotations for each task, the number of annotated images {\small $|\calI_k|$} may not be the same for different $k$. Typically the number of images with strong annotations satisfies {\small $|\calI_3| \ll n$}, as the annotation cost per sample is higher. 

The set of samples in {\small $\calI_3$} for segmentation being small, the computation of an accurate approximation of  the true probability distribution {\small $p( \bmsk{3} \!\mid\! I)$} is a challenging issue. %
Given that much more samples of $\bmsk{1}$ and $\bmsk{2}$ are observed, it is simpler to learn the statistics of these {\em weak} labels.  
Therefore, in a multi-task learning setting, one also aims at approximating  the conditional probabilities {\small $p(\bmsk{1}  \!\mid\! I)$ and  $p(\bmsk{2}  \!\mid\! I )$} for the other two tasks, or the joint probability {\small $p(\bmsk{1},\bmsk{2},\bmsk{3} \!\mid\!  I )$}.
The three tasks can be related to each other as follows. First, by the definition of the detection task, one can see that $p(\bmsk{3} \!=\! z \!\mid \!\bmsk{1}\!=\!x) = 0$ for $x$ and $z$ satisfying $x_{i,c}=1$ and $z_{i,c}=0$ for some pixel $i$ and class $c$ other than the background. Next, the map of interfaces $\bmsk{2}$ indicates small gaps between two connected instances, and is  therefore a subset of boundary pixels of the mask $\bmsk{3}$. 

Let us now consider the probabilities given by the models {\small $p(\bmsk{k} \! \mid \! I; \bmth)$ $\qut{k = 1, 2, 3}$} parameterized by {\small $\bmth$}, that will consist of network parameters in our setting. %
We do not optimize {\small $\bmth$} for individual tasks, but instead consider a joint probability {\small $p(\bmsk{1},\bmsk{2},\bmsk{3} \!\mid\!  I; \bmth )$}, so that the parameter $\bmth$ is shared among all tasks.
Assuming that {\small $\bmsk{1}$}  (rough under-segmented instance detection) and {\small $\bmsk{2}$} (a subset of shared boundaries) are conditionally independent given image {\small $I$}, 
and if the samples are i.i.d.,  we define the maximum likelihood (ML) estimator for {\small $\bmth$} as
\begin{equation}\label{eq:MLE-1}
\small
\begin{split}
{\bmth}_{\rm ML} = \arg\max_{\bmth} \sum_{I \in \calI} & \left( \log  p\qutl{ \bmsk{3} \mid \bmsk{1}, \bmsk{2}, I; \bmth} \right.  +  \sum_{k=1}^2 \left. \log p\qutl{ \bmsk{k} \mid I ; \bmth} \right).
\end{split}
\end{equation}
The set {\small $\calI_3$} may not be evenly distributed across {\small $\calI$}, but we assume that it is generated by a fixed distribution as well.  
Provided that the term {\small $\{p( \bmsk{3} \!\mid\! \bmsk{1}, \bmsk{2}, I)\}_{I \in \calI}$} can be approximated correctly by {\small $p( \bmsk{3} \!\mid\! \bmsk{1}, \bmsk{2}, I; \bmth)$} even if  {\small $\bmth$} 
is computed without {\small $\bmsk{3}$} specified for {\small $\calI \backslash \calI_3$}, then
\begin{equation}\label{eq:probability-1}
\small
\sum_{I \in \calI}   \log  p\qutl{  \!\bmsk{3}  \!\mid  \!\bmsk{1},  \!\bmsk{2},  \!I; \bmth} \! \propto \!  \sum_{I \in \calI_3}   \! \log  p\qutl{ \bmsk{3}  \!\mid  \!\bmsk{1},  \!\bmsk{2},  \!I; \bmth}.
\end{equation}
Finally assuming that the segmentation mask does not depend on {\small $\bmsk{1}$} or {\small $\bmsk{2}$} given {\small $I \in \calI_3$}, and if  {\small $|\calI_1|, |\calI_2|$} are large enough, then from Equations (\ref{eq:MLE-1}), and (\ref{eq:probability-1}),  we approximate the ML estimator by 
\begin{equation}\label{eq:MLE-2}
\small
\hat{\bmth} = \arg\max_{\bmth}  
\sum_{k=1}^{3}  \sum_{I \in \calI_k} {  \alpha_k \log  p\qutl{ \bmsk{k} \mid I; \bmth} } 
\end{equation}
in which $\alpha_1, \alpha_2, \alpha_3$ are non negative constants. 

\subsection{Loss function}
Let the outputs of the approximation models be denoted respectively by 
{\small $\hk{1}_{\bm{\theta}}(I)$, $\hk{2}_{\bm{\theta}}(I)$}, and {\small $\hk{3}_{\bm{\theta}}(I)$}, with 
{\small $\qutb{{\hk{k}_{\bm{\theta}}(I)}}_{i,c}$} the estimated probability of pixel {\small $i$} to be in class {\small $c$} of task {\small $k$}.
For each task $k$, %
the log likelihood function related to  the label {\small $\bmsk{k}$} writes
\begin{equation}\label{logP}
\small
\log  p\qutl{ \bmsk{k} \mid I; \bmth}\!
= \! \sum_{i} \! \sum_{c \in C_k}\! \sk{k}_{i,c} \log \qutb{{\hk{k}_{\bm{\theta}}(I)}}_{i,c}, \ \ k = 1, 2, 3,
\end{equation}
in which {\small $\sk{k}_{i,c}$} denotes the element of the label {\small $\bmsk{k}$} at pixel {\small $i$} for class {\small $c$} and {\small $C_k$} is the set of classes for task {\small $k$}. For example, for SEM images of ice cream (see details in section \ref{sec:SEM}), we have three classes including air bubbles, ice crystals and the rest (background or parts of the objects ignored by the weak labels), so {\small $C_1, C_3 = \left\{1,2,3\right\}$}. For the separation task, there are only two classes for pixels (belonging or not to a touching  interface) and {\small $C_2 = \left\{1,2\right\}$}.  According to Equation (\ref{eq:MLE-2}), the network is trained by minimizing the weighted cross entropy loss: %
\begin{equation}\label{eq:loss}
\small
L\qut{\bmth} = - \sum_{I \in \calI} \sum_{k=1}^{3} \alpha_k  \mathbbm{1}_{\calI_k}
\qutl{I}\log  p\qutl{ \bmsk{k} \mid I; \bmth},%
\end{equation}
Here {\small $ \mathbbm{1}_{\calI_k}\qutl{\cdot}$} is an indicator function which is $1$ if {\small $I \in \calI_k$} and {\small $0$} otherwise. 

\subsection{Multi-task Network}
\begin{figure}
  \centering
\includegraphics[width=\linewidth, trim=5  197 0 151 ,clip]{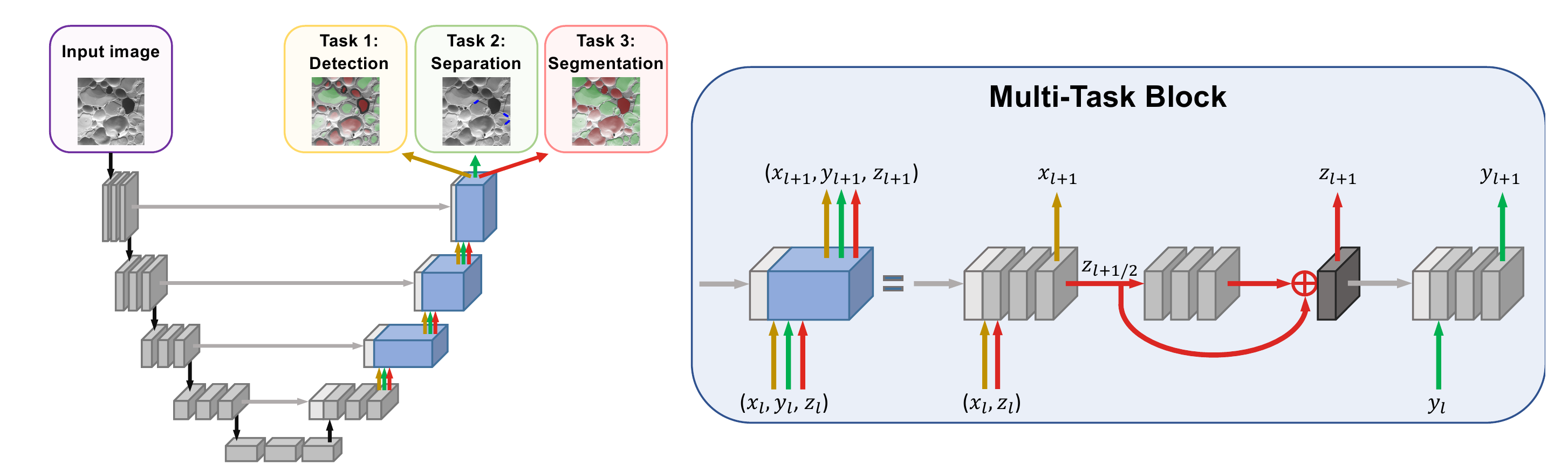}
  \caption{Architecture of the multi-task U-net. 
The left part of the network is a contracting path similar to the standard U-net. %
For multi-task learning, we construct several expansive paths with specific multi-task blocks.
At each resolution, task 1 (Detection in yellow) and task 3 (Segmentation in red) run through a common sub-block, 
but the red path learns an additional residual to better localize object boundaries. 
Long skip connections with the layers from contracting path are built for yellow/red paths via concatenation. 
Task 2 (Separation, in green) mainly follows a separated expansive path, with its own up-sampled blocks. 
A link with the last layer of task 3 is added via a skip connection in order to integrate accurate  boundaries in the separation task. 
}\label{fig:network-arch}
\end{figure}

We follow a convolutional encoder-decoder network structure for multi-task learning. 
The network architecture is illustrated in Figure \ref{fig:network-arch}.
As an extension of the U-net structure for multiple tasks,
we only have  one contracting path that encodes shared features representation for all the tasks. 
On the expansive branch, we introduce a multi-task block at each resolution to support different learning purposes (blue blocks in Figure \ref{fig:network-arch}). 
Every multi-task block runs three paths, with three inputs and three corresponding outputs, 
and it consists of several sub-blocks.%

In each multi-task block, 
the detection task (task 1) and the segmentation task (task 3) have a common path similar to the decoder part of the standard U-net. 
They share the same weights and use the same concatenation with feature maps from contracting path via the skip connections.
However, we insert an additional residual sub-block for the segmentation task. 
The residual sub-block provides extra network parameters to learn information not known from the detection task, \textit{e.g.} object boundary localization.
The path for the separation task (task 2) is built on the top of detection/segmentation ones. 
It is also a U-net decoder block structure, but the long skip connections start from the sub-blocks of the detection/segmentation paths instead of the contracting path.  The connections extract higher resolution features from the segmentation task and use them in the separation task.

To formulate the multi-task blocks, 
let {\small $\bmx_l$} and {\small $\bmz_l$} denote respectively the output of the detection path and segmentation path at the multi-task block {\small $l$},
and let $\bmc_l$ be the feature maps received from the contracting path with the skip connections. 
Then for task 1 and task 3 we have  
\begin{equation}\label{eq:block_farward}
\small
    \begin{cases}
& \bmx_{l+1} \!=\! F_{W_l}(\bmx_l, \bmc_l), \\
& \bmz_{l+\frac{1}{2}} \!=\! F_{W_l}(\bmz_l, \bmc_l), \quad
\bmz_{l+1} \!=\! \bmz_{l+\frac{1}{2}} \!+\! F_{W_{l+\frac{1}{2}}}\!(\bmz_{l+\frac{1}{2}}),    
    \end{cases}
\end{equation}
in which {\small $W_{l}, W_{l+1/2} \in \bm{\theta}$} are subsets of network parameters and $F_{W_l}$, $F_{W_{l+\frac{1}{2}}}$ are  respectively determined by a sequence of layers of the network (cf.  small grey blocks on the right of Figure \ref{fig:network-arch}).
For task 2 the output at {\small $l^{\rm th}$} block {\small $\bmy_{l+1}$} is computed as 
{\small $\bmy_{l+1} = G_{\tilde{W_l}}\qut{\bmz_{l+1}, \bmy_{l}}$}
with additional network parameters {\small $\tilde{W_l} \in \bm{\theta}$}.  Finally, after the last multi-task block, softmax layers are added, outputting a probability map for each task.

\noindent{\bf Implementation details.}
We implement a multi-task U-net with $6$ levels of spatial resolution and input images of size {\small $256 \times 256$}. 
A sequence of down-sampling via max-pooling with pooling size {\small $2\times 2$} is used for the contracting path of the network. 
Different from the conventional U-net \cite{ronneberger2015u}, 
each small grey block (see Figure \ref{fig:network-arch}) consists of a convolution layer and a batch normalization \cite{ioffe2015batch}, followed by a leaky ReLU activation with a leakiness parameter {\small $0.01$}. 
The same setting is also applied to grey sub-blocks of the {\small $4$} multi-task blocks. 
On the expansive path of the network, feature maps are up-sampled (with factor {\small $2\times 2$}) by bilinear interpolation from a low resolution multi-task block to the next one. 

\subsection{Methods for lazy labels generation}\label{subs:labelgenerate}
We now explain our strategy for generating all the lazy annotations that are used for training. 
We introduce our method with a data set of ice cream SEM images but any other similar microscopy datasets could be used. Typical images of ice cream samples are shown in the top row of the left part of  Figure \ref{tab:annotated_images}. The segmentation problem is challenging since the images contain densely distributed small object instances (\textit{i.e.}, air bubble and ice crystals), and poor contrast between the foreground and the background.  The sizes of the objects can vary significantly in a single sample. Textures on the surfaces of objects also appear. 

As a first step,  scribble-based labelling is applied to obtain detection regions of air bubbles and ice crystals for task 1. 
This can be done in a very fast way as no effort is put on the exact object boundaries. 
We adopt a lazy strategy by picking out an inner region for each object in the images (see \textit{e.g.}, the second row of the left part of Figure \ref{tab:annotated_images}).  
Though one could get these rough regions as accurate as possible, we delay such refinement to task 3, for better efficiency of the global annotation process. %
Compared to the commonly used bounding box annotations in computer vision tasks, these labels give more confidence for a particular part of the region of interest. 

In the second step, we focus on tailored labels for those instances that are close one to each other (task 2), without a clear boundary separating them.  
Again, we use scribbles to mark their interface. %
Examples for such annotations are given in Figure \ref{tab:annotated_images} (top line, right part).
The work can be carried out efficiently especially when the target scribbles have a sparse distribution. 
Lazy manual labelling of tasks 1 and 2 are done independently. It follows the assumption made in Section~\ref{sec:model}
that $\bmsk{1}$  and $\bmsk{2}$ are conditionally independent given image $I$.

The precise labels for task $3$ are created using interactive segmentation tools. %
We use Grabcut \cite{rother2004grabcut,} a graph-cut based method. The initial labels obtained from the first step give a good guess of the whole object regions. 
The Grabcut works well on isolated objects. However, it gives poor results when the objects are close to each other and  have boundaries with inhomogeneous colors. %
As corrections may be needed for each image,  only a few images of the whole dataset  are processed.
A fully segmented example is shown in the last row of Figure \ref{tab:annotated_images}.

\section{Experiments}\label{sec:exp}
In this section, we demonstrate the performance of our approach using two microscopy image datasets. For both, we use strong labels (SL) and weak labels (WL).  
We prepare the labels and design the network as described in Section \ref{sec:model}. 
\subsection{Segmenting SEM images of ice cream}\label{sec:SEM}
Scanning Electron Microscopy (SEM) constitutes the state-of-the-art for analysing food microstructures as it 
enables the efficient acquisition of high quality images for food materials, 
resulting into a huge amount of image data available for analysis. 
However, to better delineate the microstructures and provide exact statistical information, 
the partition of the images into different structural components and instances is needed. 
The structures of food, especially soft solid materials, are usually complex which makes automated segmentation a difficult task. 
Some SEM images of ice cream in our dataset are shown on the bottom right of Figure \ref{fig:icecreamimages}. 
A typical ice cream sample consists of air bubbles, ice crystals and a concentrated unfrozen solution. 
We treat the solution as the background and aim at detecting and computing a pixel-wise classification for each air bubbles and ice crystals instances.

\renewcommand{\sidecap}[1]{ \raisebox{-.4\height}{\begin{sideways}\parbox{0.18\linewidth}{\centering  #1}\end{sideways}} }
\begin{figure}[h!]
	\centering
 	\begin{minipage}[b]{0.48\linewidth}
		\begin{tabular}{@{\hskip-5pt}c@{\hskip1pt}m{1.1cm}@{\hskip1pt}m{1.1cm}@{\hskip1pt}m{1.1cm}@{\hskip1pt}m{1.1cm}@{\hskip1pt}m{1.1cm}@{\hskip1pt}}
		\sidecap{\scriptsize Images} &
		\includegraphics[width=0.98\linewidth]{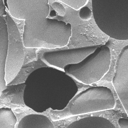} &
		\includegraphics[width=0.98\linewidth]{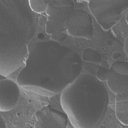} &
		\includegraphics[width=0.98\linewidth]{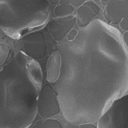} &
		\includegraphics[width=0.98\linewidth]{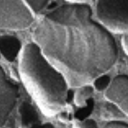} &
		\includegraphics[width=0.98\linewidth]{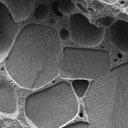} %
		\\
		\sidecap{\scriptsize Task 1} &
		\includegraphics[width=0.98\linewidth]{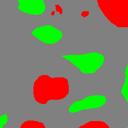} &
		\includegraphics[width=0.98\linewidth]{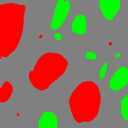} &
		\includegraphics[width=0.98\linewidth]{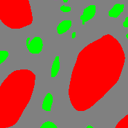} &
		\includegraphics[width=0.98\linewidth]{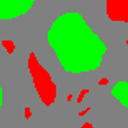} & \\
\end{tabular}
\end{minipage}
\hfill\vline\hfill
\begin{minipage}[b]{0.48\linewidth}
\begin{tabular}{@{\hskip-1pt}c@{\hskip3pt}m{1.1cm}@{\hskip1pt}m{1.1cm}@{\hskip1pt}m{1.1cm}@{\hskip1pt}m{1.1cm}@{\hskip1pt}m{1.1cm}@{\hskip1pt}}
	\sidecap{\scriptsize Task 2}&
	\includegraphics[width=0.98\linewidth]{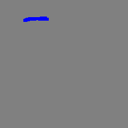} &
	\includegraphics[width=0.98\linewidth]{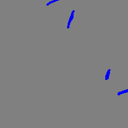} &
	\includegraphics[width=0.98\linewidth]{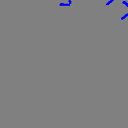} &
	\includegraphics[width=0.98\linewidth]{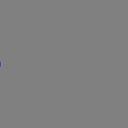} &
	\includegraphics[width=0.98\linewidth]{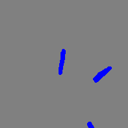}%
	\\
	\sidecap{\scriptsize Task 3}&
	\includegraphics[width=0.98\linewidth]{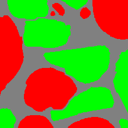} 
	& & & & %
\end{tabular}
	\end{minipage}
	\caption{Example of annotated images. %
		Some of the annotations are not shown because 
		the images are not labelled for the associated tasks. 
		The red color and green color are for air bubbles and ice crystals, respectively. 
		The blue curves in Task 2 are labels for interfaces of touching objects. 
	}\label{tab:annotated_images}
\end{figure}

The set of ice-cream SEM dataset consists of $38$ wide field-of-view and high resolution images that are split into three sets (53\% for training, 16\% for validation and 31\% testing respectively). Each image contains a rich set of instances with an overall number of instances around $13300$ for $2$ classes (ice crystals and air bubbles). For comparison,  the PASCAL VOC 2012 dataset has $27450$ objects in total for $20$ classes.

~\\
\begin{minipage}{\textwidth}
\centering
\quad 
\begin{minipage}[c]{0.45\textwidth}
	\centering
	\tabcaption{Dice scores of segmentation results on the test images of SEM images of ice cream dataset.
    }\label{tab:single-task}
    \makebox[\textwidth][c]{
	\begin{tabular}{c|ccc}
		\toprule
		The models  & {\scriptsize air bubbles} & {\scriptsize ice crystals} & Overall \\
		\hline
		U-net on WL & 0.725& 0.706 & 0.716 \\
		U-net on SL & 0.837 & 0.794 & 0.818 \\
		PL approach & 0.938 & 0.909 & 0.924 \\
		Multi-task U-net &  \textbf{0.953}& \textbf{0.931} & \textbf{0.944} \\
		\bottomrule
	\end{tabular}}
\end{minipage}
\quad \quad 
\begin{minipage}[c]{0.45\textwidth}
    \centering
    \includegraphics[width=0.7\linewidth, trim=20 15 20 20, clip]{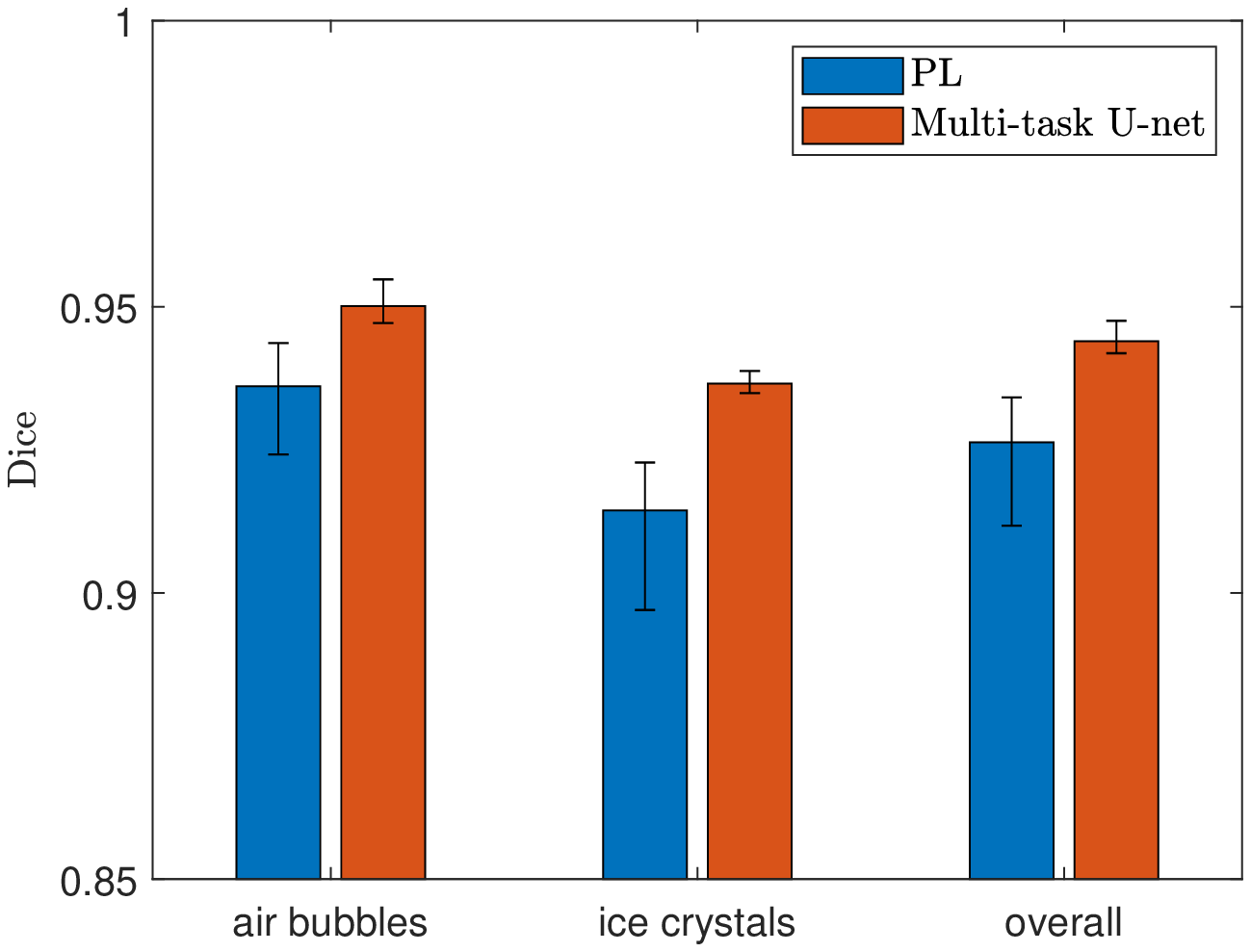}
    \figcaption{The error bars for the PL and multi-task U-net. The top of each box represent the mean of the scores over $8$ different experiments, the minimum and maximum of which are indicated by the whiskers }\label{icecream_errorbar}
\end{minipage}
\end{minipage}
~\\

For training the network, data augmentation is applied to prevent over-fitting.
The size of the raw images is $960 \times 1280$. They are rescaled and rotated randomly, and then cropped into an input size of $256 \times 256$ for feeding the network.
Random flipping is also performed during training. The network is trained using Adam optimizer \cite{kingma2014adam} with a learning rate $r=2\times 10^{-4}$ and a batch size of $16$. 

In the inference phase, the network outputs for each patch a probability map of size $256 \times 256$. The patches are then aggregated to obtain a probability map for the whole image. In general, the pixels near the boundaries of each patch are harder to classify. We thus weight the spatial influence of the patches with a Gaussian kernel to emphasize the network prediction at patch center.

We now evaluate the multi-task U-net and compare it to the traditional single task U-net. %
The performance of each model is tested on 12 wide FoV images, and average results are shown in Table \ref{tab:single-task}. 
In the table, the dice score for a class {\small $c$} is defined as 
{\small 
$
d_c = { 2 \sum_{i} x_{i,c} y_{i,c}   }/\qutl{  \sum_i x_{i,c} + \sum_i y_{i,c} }
$}%
where  {$x$} is the computed segmentation mask  and $y$ the ground truth.

~\\
\definecolor{atomictangerine}{rgb}{1.0, 0.6, 0.4}
\definecolor{cyanprocess}{rgb}{0.0, 0.72, 0.92}
\def\ecws{1.125}
    \begin{minipage}{\linewidth}
    \centering
	\begin{minipage}[b]{0.48\linewidth}
	    \setlength{\tabcolsep}{1pt}
		\centering
		\begin{tabular}{cc}
            \begin{tikzpicture}[spy using outlines={magnification=1.8,size=30*\ecws}]
    	  \node{\includegraphics[width=2.7cm]{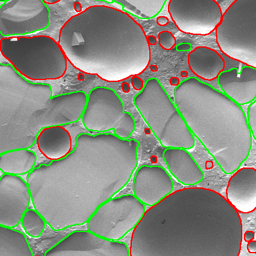}};  
    	  \spy [atomictangerine,line width=30] on (0.225*\ecws,0.825*\ecws) in node at (0.65*\ecws,-1.8*\ecws);
    	    \spy [cyanprocess, line width=30] on (0,-0.9*\ecws) in node at (-0.65*\ecws,-1.8*\ecws);
            \end{tikzpicture} 
			&
			\begin{tikzpicture}[spy using outlines={magnification=1.8,size=30*\ecws}]
			    \node{\includegraphics[width=2.7cm]{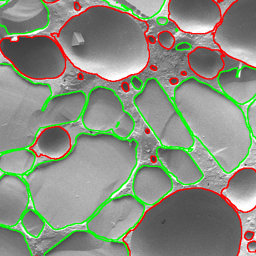}};
        	    \spy [atomictangerine,line width=30] on (0.225*\ecws,0.825*\ecws) in node at (0.65*\ecws,-1.8*\ecws);;
        	    \spy [cyanprocess, line width=30] on (0,-0.9*\ecws) in node at (-0.65*\ecws,-1.8*\ecws);;
            \end{tikzpicture}
			\\
			\begin{tikzpicture}[spy using outlines={magnification=1.8,size=30*\ecws}]
			\node{\includegraphics[width=2.7cm]{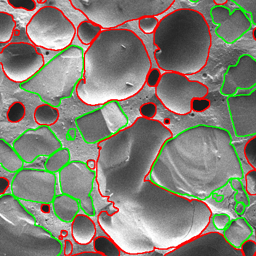}};
			    \spy [atomictangerine,line width=30] on (0.225*\ecws,0.675*\ecws) in node at (0.65*\ecws,-1.8*\ecws);;
			    \spy [cyanprocess, line width=30] on (-0.9*\ecws,0.825*\ecws) in node at (-0.65*\ecws,-1.8*\ecws);;
            \end{tikzpicture}
			&
			\begin{tikzpicture}[spy using outlines={magnification=1.8,size=30*\ecws}]
			\node{\includegraphics[width=2.7cm]{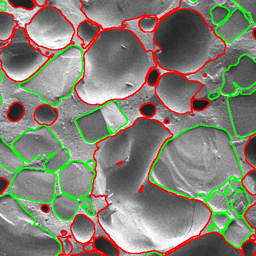}};
			    \spy [atomictangerine,line width=30] on (0.225*\ecws,0.675*\ecws) in node at (0.65*\ecws,-1.8*\ecws);;
			    \spy [cyanprocess, line width=30] on (-0.9*\ecws,0.825*\ecws) in node at (-0.65*\ecws,-1.8*\ecws);;
            \end{tikzpicture} 
			\\
			{\small multi-task U-net}&PL%
		\end{tabular}
	\end{minipage}
	\quad  %
	\vline
	\begin{minipage}[b]{0.48\linewidth}
	    \setlength{\tabcolsep}{1pt}
		\centering
		\begin{tabular}{@{\hskip5pt}cc}
			\begin{tikzpicture}[spy using outlines={magnification=1.8,size=30*\ecws}]
			   \node{\includegraphics[width=2.7cm,trim=0 0 160 0 ,clip]{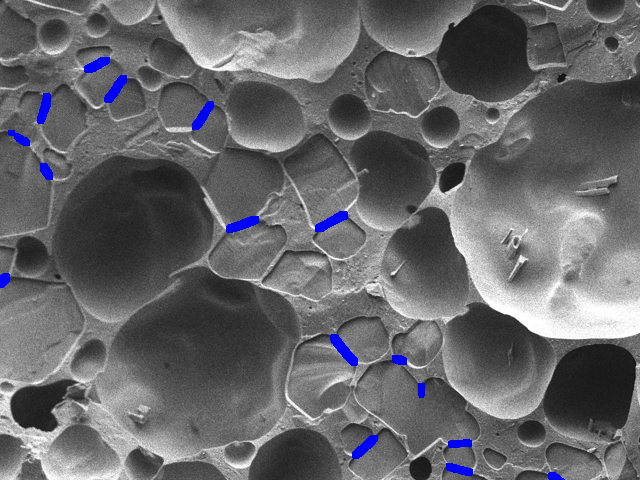}};
			    \spy [atomictangerine,line width=30] on (-0.6*\ecws,0.75*\ecws) in node at (-0.65*\ecws,-1.8*\ecws);
			\end{tikzpicture} 
			&
			\begin{tikzpicture}[spy using outlines={magnification=1.8,size=30*\ecws}]
			  \node{\includegraphics[width=2.7cm,trim=0 0 160 0 ,clip]{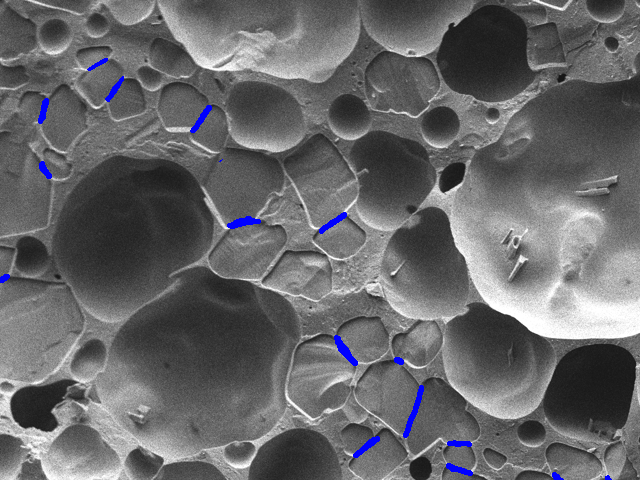}};
			    \spy [atomictangerine,line width=30] on (-0.6*\ecws,0.75*\ecws) in node at (-0.65*\ecws,-1.8*\ecws);
			\end{tikzpicture}
			\\
			\begin{tikzpicture}[spy using outlines={magnification=1.8,size=30*\ecws}]
			   \node{\includegraphics[width=2.7cm,trim=0 0 160 0 ,clip]{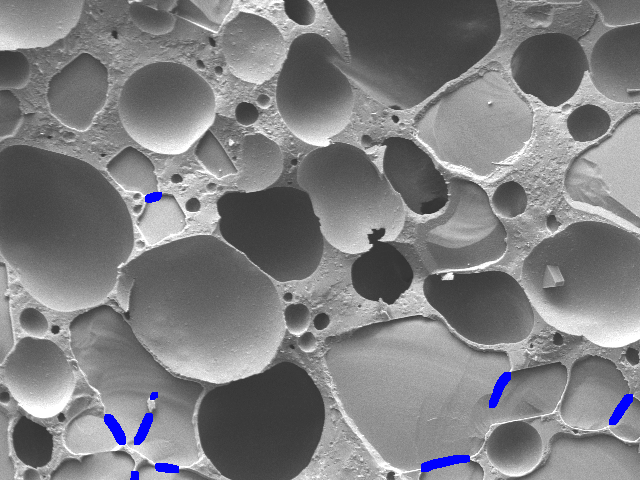}};
			    \spy [atomictangerine,line width=30] on (-0.45*\ecws,-0.9*\ecws) in node at (-0.65*\ecws,-1.8*\ecws);
			\end{tikzpicture} 
			&
			\begin{tikzpicture}[spy using outlines={magnification=1.8,size=30*\ecws}]
			   \node{\includegraphics[width=2.7cm,trim=0 0 160 0 ,clip]{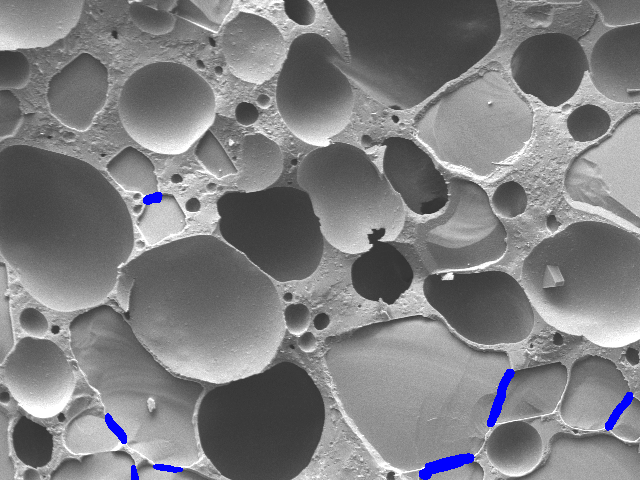}};
			    \spy [atomictangerine,line width=30] on (-0.45*\ecws,-0.9*\ecws) in node at (-0.65*\ecws,-1.8*\ecws);
			 \end{tikzpicture} 
			\\
			{\small \begin{tabular}{c}
			multi-task U-net \\ separation
			\end{tabular}}&{\small true separation}
		\end{tabular}
	\end{minipage}
	\figcaption{Segmentation and separation results (best view in color). First two columns: the computed contours are shown in red for air bubbles and green for ice crystals. While multi-task U-net and PL supervised network both have good performance, PL misclassifies the background near object boundaries. Last two columns: Examples of separation by the multi-task U-net and the ground truth. }\label{tab:multi-taskU-SDI}
\end{minipage}
~\\

We train a single task U-net (\textit{i.e.}, without the multi-task block) on the weakly labelled set (task 1), with the $15$ annotated images.  %
The single task U-net on weak annotations gives an overall dice score at $0.72$, 
the lowest one among the three other methods tested.  
One reason for the low accuracy of the single task U-net on weak (inaccurate) annotations is that in the training labels,  
the object boundaries are mostly ignored. Hence the U-net is not trained to recover them, 
leaving large parts of the object not recognized. 
Second, we consider strong annotations as training data, without the data of the other tasks, \textit{i.e.} only 2 images with accurate segmentation masks are used. The score of the U-net trained on SL is only  $0.82$, which is  significantly lower than the $0.94$ obtained by our multi-task network.  

We also compare our multi-task U-net results with one of the major weakly supervised approaches that make use of pseudo labels (PL) (see e.g., \cite{khoreva2017simple, jing2018coarse}).
In these approaches, the pseudo segmentation masks are created from WLs and are used to feed a segmentation network.
Following the work of \cite{khoreva2017simple}, we use the Grabcut method to create the PLs from the partial masks of task 1. For the small subset of images that are strongly annotated, the full segmentation masks are used instead of PLs. 
The PLs are created without human correction, and then used for feeding the segmentation network. Here we use the single task U-net for baseline comparisons. %

Our multi-task network outperforms the PL approach as shown in Table \ref{tab:single-task}. Figure \ref{icecream_errorbar} displays the error bars for the two methods with dice scores collected from $8$ different runs. 
The performance of the PL method relies on the tools used for pseudo segmentation mask generations. If the tools create bias in the pseudo labels, then the learning will be biased as well, which is the case in this example. 
The images in the left part of Figure \ref{tab:multi-taskU-SDI} show that the predicted label of an object tends to merge with some background pixels when there are edges of another object nearby.

Besides the number of pixels that are correctly classified, the separation of touching instances is also of interest. 
In addition to the dice scores in Table \ref{tab:single-task}, we study the learning performance of our multi-task network on task 2, which specializes in the separation aspect. 
The test results on the $12$ images give an overall precision of $0.70$ of the detected interfaces, while $0.82$ of the touching objects are recognized. We show some examples of computed separations and ground truth in the right part of Figure \ref{tab:multi-taskU-SDI}.

\begin{figure}[h]
\setlength\tabcolsep{5pt}
\small
\centering
    \begin{tabular}{ccc}
        \includegraphics[width=0.28\linewidth, angle=90]{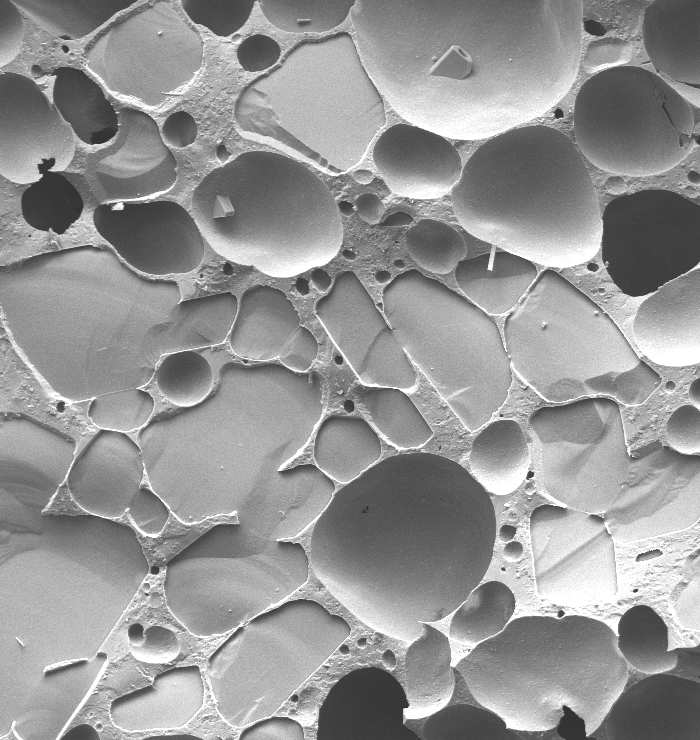}&
        \includegraphics[width=0.28\linewidth, angle=90]{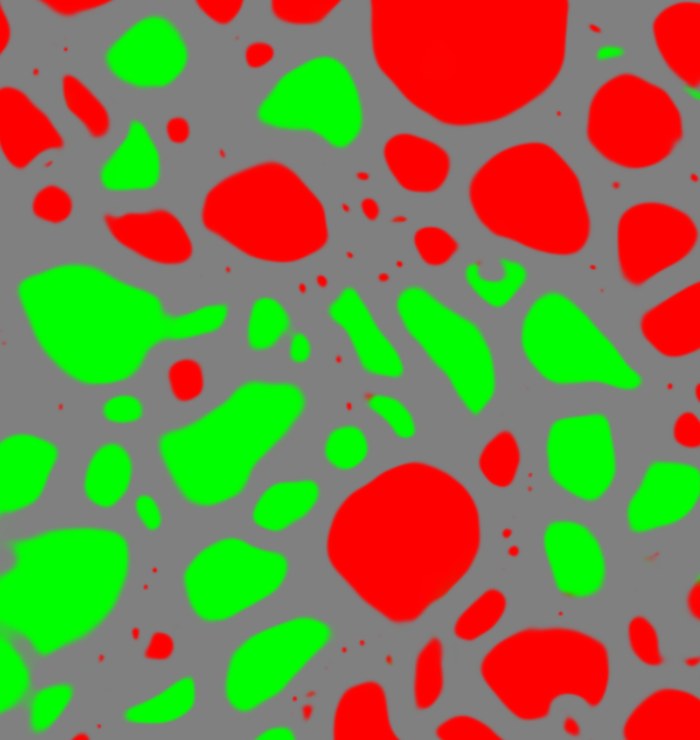}&
        \includegraphics[width=0.28\linewidth, angle=90]{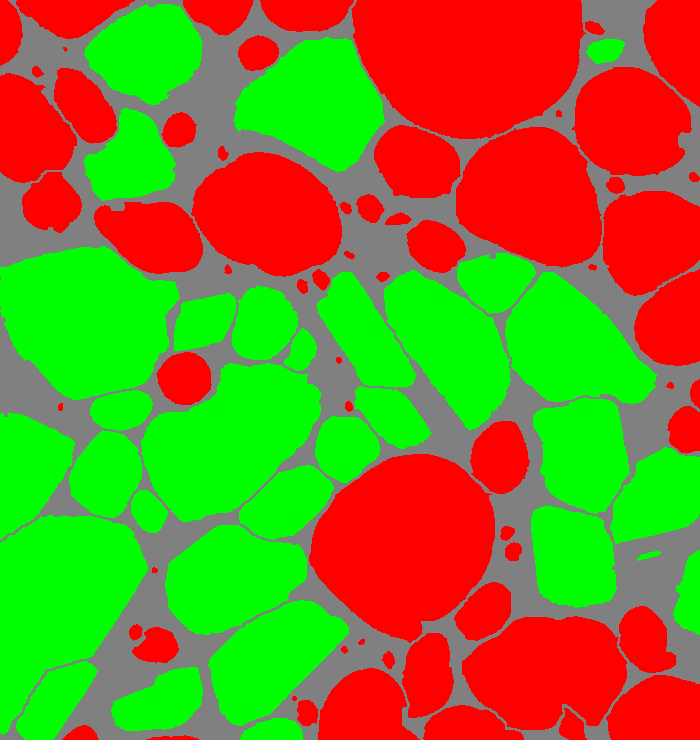}         
    \end{tabular}
    \caption{The image (left), the inaccurate label predicted by the network for the detection task (middle), and the ground truth segmentation mask (right). The {red} and {green} colors on the middle and right images stand for air bubbles and ice crystals respectively.}\label{tab:R6}
\end{figure}
For the detection task, the network predicts a probability map for the inner regions of the object instances. An output of the network is shown in Figure \ref{tab:R6}. With partial masks as coarse labels for this task, the network learns to identify the object instances. %

\subsection{Gland segmentation on H\&E-stained images}
We apply the approach to the segmentation of tissues in histology images. In this experiment, we use the GlaS challenge dataset \cite{sirinukunwattana2017gland} that consists of $165$ Hematoxylin and eosin (H\&E) stained images. The dataset is split into three parts, with $85$ images for training, and $60$ for offsite test and $20$ images for onsite test (we will call the latter two sets Test part A and Test part B respectively in the following). 

Apart from the SL available from the dataset, we create a set of a weak labels for the detection task and separation task.
These weak labels together with a part of the strong labels are used for training the multi-task U-net.

\begin{table}
    \centering
    \setlength{\tabcolsep}{4pt}
    \caption{Average dice score for segmentation of gland. 
    Results of two sets of methods, weakly supervised (WS) and strongly supervised (SS) are displayed.
    Our method uses both SL and WL. The ratio of strong labels (SL) is increased from $2.4\%$ to $100\%$, and the scores of the methods are reported here for two parts A and B of the test sets, as split in  \cite{sirinukunwattana2017gland}.
    }\label{tab:R2}
       \small
    \begin{tabular}{|c|c|c|c|c|c|c|}
        \toprule \multicolumn{3}{|c|}{SL Ratio} & 2.4\% & 4.7\% & 9.4\% & 100\%\\
         \hline
         \multirow{4}{*}{\makecell{Test\\ Part A}} &  \multirow{3}{*}{WS} &
         Ours & \textbf{0.866} & \textbf{0.889} & \textbf{0.915} & \textbf{0.921} \\
         &&  Single task & 0.700 & 0.749 & 0.840 & 0.921 \\
         && PL & 0.799 & 0.812 & 0.820 & \\
         \cline{2-7}
         & SS & MDUnet & & & & 0.920 \\
         \hline
         \multirow{4}{*}{\makecell{Test\\ Part B}} & \multirow{3}{*}{WS} &
         Ours & \textbf{0.751} & \textbf{0.872} & \textbf{0.904} & \textbf{0.910} \\
         && Single task & 0.658 & 0.766 & 0.824 & 0.908 \\
         && PL & 0.773 & 0.770 & 0.782 & \\
         \cline{2-7}
         & SS & MDUnet & & & & 0.871 \\
         \bottomrule
    \end{tabular}
\end{table}

\begin{figure}[t!]
\centering
\setlength\tabcolsep{20pt}
\begin{tabular}{cc}
	\includegraphics[width=0.35\linewidth]{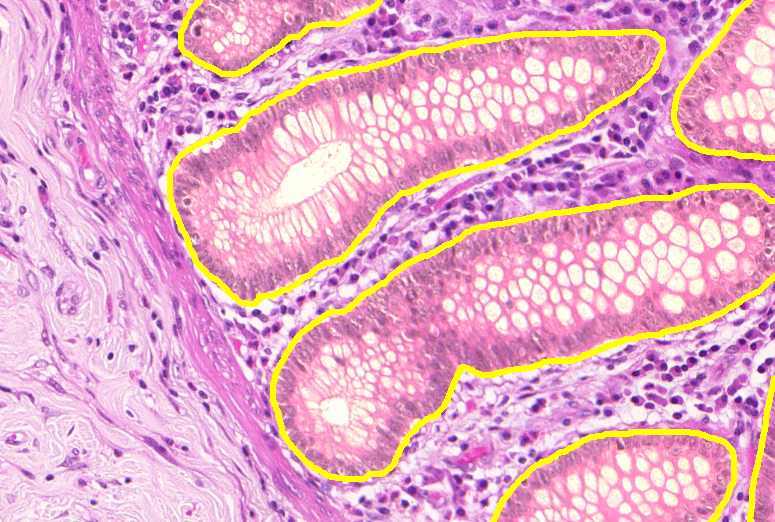} & 
	\includegraphics[width=0.35\linewidth]{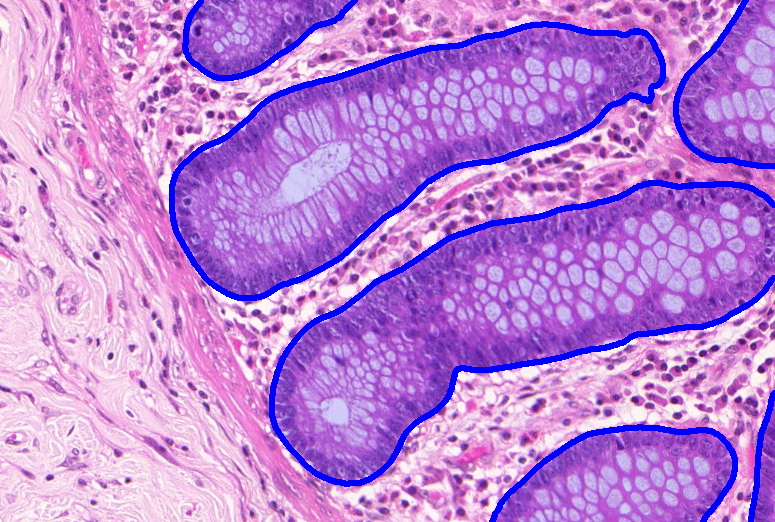} \\ 
	\addlinespace[-0.5ex]
	(a) ground truth & (b) U-net (100 \% SL) \\
	\addlinespace[1.5ex]
	\includegraphics[width=0.35\linewidth]{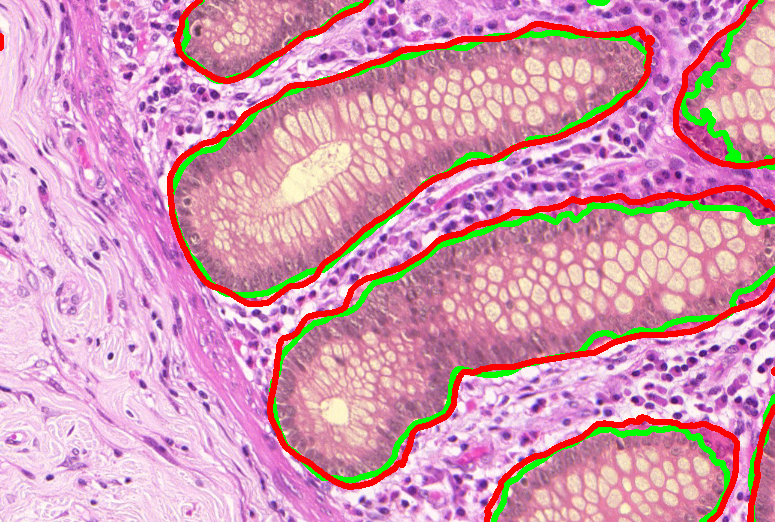} & 
	\includegraphics[width=0.35\linewidth]{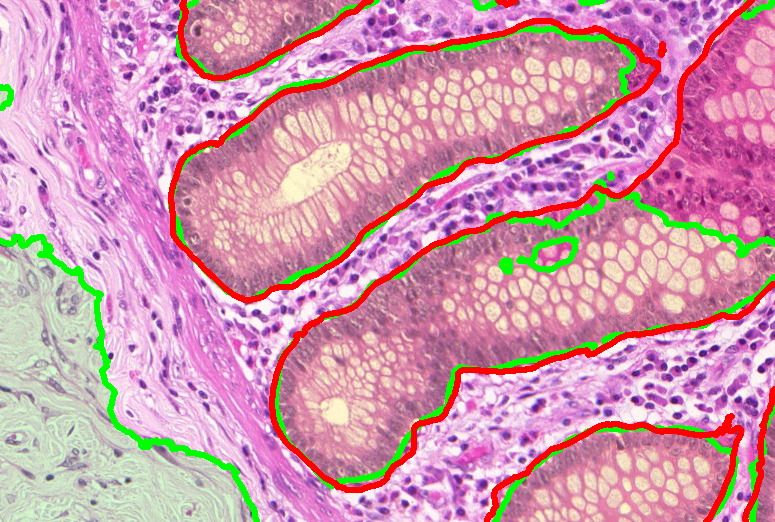} 	\\
	\addlinespace[-0.5ex]
	\makecell{ (c) multi-task\\ {\tiny ({Red}: \textbf{$9.4\%$} SL,  {Green}: \textbf{$4.7\%$} SL)}}  & \makecell{(d) U-net\\ ({\tiny {Red}: \textbf{$9.4\%$} SL, {Green}: \textbf{$4.7\%$} SL)}}
\end{tabular}
\caption{Segmentation results on the gland dataset (best view in color). The {ground truth} and the results. For (c) and (d), {Red} contour denotes the results from \textbf{$9.4\%$} strong labels; {Green} contour denotes results from \textbf{$4.7\%$} strong labels.}\label{fig:gland_segmented_images}
\end{figure}

In this experiment, we test the algorithm on different ratios of SL, and compare it with the baseline U-net (single task), PL approach (where PL are generated in the same way as the ones for the SEM dataset), and a fully supervised approach called Multi-scale Densely Connected U-Net (MDUnet) \cite{zhang2018mdu}. The results on two sets of test data are reported in Table \ref{tab:R2}. As the SL ratios increase from $2.4\%$ to $9.4\%$, an improvement of performance of the multi-task U-net is gained. When it reaches $9.4\%$ SL, the multi-task framework achieves comparable score with the fully supervised version, and outperforms the PL approach by a significant margin. We emphasize that the $9.4\%$ SL and WL 
can be obtained several times faster than
the $100\%$ SL used for fully supervised learning. Example of segmentation results are displayed in Figure \ref{fig:gland_segmented_images}.

\section{Conclusion}

In this paper, we develop a multi-task learning framework for microscopy image segmentation, which relaxes the requirement for numerous and accurate annotations to train the network. It is therefore suitable for  segmentation problem with a dense population of object instances.
The model separates the segmentation problem into three smaller tasks. 
One of them is dedicated  to the instance detection and therefore does not need exact boundary information. 
This gives potential flexibility as one could concentrate on the classification and rough location of the instances during data collection. The second one focuses on the separation of objects sharing a common boundary.
The final task aims at extracting pixel-wise boundary information.  
Thanks to the information shared within the multi-task learning, this accurate segmentation can be obtained using very few annotated data.
Our model is end-to-end and requires no reprocessing for the weak labels. 
For the partial masks that ignore boundary pixels, the annotation can also be done when the boundaries of object are hard to detect.
In the future, we could like to extend the proposed approach for solving 3D segmentation problems in biomedical images where labelling a single 3D image needs much more manual work.\\ 

\noindent\textbf{Acknowledgments}
RK and CBS acknowledge support from the EPSRC grant EP/T003553/1. CBS additionally acknowledges support from the Leverhulme Trust project on `Breaking the non-convexity barrier', the Philip Leverhulme Prize, the EPSRC grant EP/S026045/1, the EPSRC Centre Nr. EP/N014588/1, the RISE projects CHiPS and NoMADS, the Cantab Capital Institute for the Mathematics of Information and the Alan Turing Institute, Royal Society Wolfson fellowship. AB and NP acknowledge support from the EU Horizon 2020 research and innovation programme NoMADS (Marie Sk\l{}odowska-Curie grant agreement No. 777826).

\bibliographystyle{splncs04}
\bibliography{DL_segmentation_bio_ref}
\end{document}